\documentclass[letterpaper,twocolumn,10pt]{article}
\usepackage{usenix2020_SOUPS}

\usepackage{tikz}
\usepackage{amsmath}

\usepackage{float}

\usepackage{filecontents}

\begin{document}

\date{}

\title{\Large \bf LLM-Based Assistant for Dog Care Guidance\\}

\def\plainauthor{Author name(s) for PDF metadata. Don't forget to anonymize for submission!}

\author{
{\rm Abubakar Sadiq Shittu}\\
University of Tennessee, Knoxville
\and
{\rm Venkat Gopu}\\
University of Tennessee, Knoxville
\and
{\rm Arshita Sharma}\\
University of Tennessee, Knoxville
\and
{\rm Jacob Kennedy}\\
University of Tennessee, Knoxville
}

\maketitle

\begin{abstract}

This study investigates whether domain grounding through Retrieval-Augmented Generation (RAG) and parameter-efficient fine-tuning using Low-Rank Adaptation (LoRA) can enhance the accuracy and reliability of pet care advice generated by large language models (LLMs). We curated a dataset of breed-specific dog care information and used it to develop two systems: a RAG pipeline based on TinyLlama-1.1B-Chat-v1.0, and a fine-tuned LLaMA-3.2-1B model adapted using LoRA. Both systems were evaluated on 15 dog care questions. The RAG model achieved a perfect average score of 2.0, outperforming the fine-tuned model (mean = 1.47). Statistical analyzes, including Welch's t test, Mann-Whitney U, and Wilcoxon signed rank tests (all p < 0.05), confirmed the significance of the performance gap. Our findings suggest that retrieval-based domain grounding leads to more accurate, consistent, and trustworthy responses than standalone fine-tuning in specialized LLM applications.

\end{abstract}

\section{Introduction}

Large Language Models (LLMs) have recently surged in popularity due to their ability to generate fluent, human-like responses across a wide range of general-purpose tasks, including summarization, translation, and open-domain question answering~\cite{c6, c7}. While these models demonstrate impressive performance on broad benchmarks, they often struggle when applied to domain-specific tasks that require precise, context-sensitive reasoning and factual accuracy~\cite{c7}. One such domain is breed-specific dog care, where small distinctions between breeds can significantly impact health, nutrition, behavior, and grooming practices.

Pet owners commonly rely on digital assistants or search engines to obtain timely guidance on caring for their dogs. However, unlike typical consumer queries, dog care questions require highly specialized responses that account for breed, size, age, health predispositions, and lifestyle factors. For example, an ideal feeding plan for an energetic Border Collie may be inappropriate for a senior Shih Tzu with dental issues. Despite these nuances, off-the-shelf LLMs often return generic or overly broad responses, sometimes even inconsistent or incorrect, due to their lack of grounding in verified veterinary knowledge. These models frequently hedge with disclaimers such as ``consult your veterinarian,'' underscoring their limitations in delivering concrete, evidence-based advice.

To address these shortcomings, we explore two central research questions:

\begin{enumerate}
    \item Can domain grounding improve the factual accuracy and reliability of pet care advice generated by LLMs?
    \item What is the performance gap between default LLM outputs and responses enhanced via Retrieval-Augmented Generation (RAG) or fine-tuning?
\end{enumerate}

These questions are crucial, as pet owners depend on precise, breed-specific guidance to ensure the well-being of their dogs. Answers to these questions could directly influence care practices, improve animal health outcomes, and bolster user confidence in AI-generated advice.

To this end, we propose an approach that integrates both RAG and parameter-efficient fine-tuning using Low-Rank Adaptation (LoRA) on a curated dataset of verified dog breed information. This method is designed to overcome the limitations of ungrounded LLMs by injecting accurate and relevant domain knowledge into the model’s response generation process. Through this dual approach, we aim to evaluate the effectiveness of domain-grounded LLMs in delivering high-quality, trustworthy pet care guidance.

\section{Related work}
RAG has emerged as a popular method for improving the domain specificity and factual grounding of LLM.

A closely related work is by Sun et al. \cite{c1}, who developed a RAG-based QA system focused on Pittsburgh and Carnegie Mellon University. Their pipeline, which combines BM25 and FAISS with a reranker, significantly improved exact match and F1 scores compared to baseline LLMs. However, their training data relied on synthetic question-answer pairs generated by language models and noisy web-scraped sources. While effective in boosting performance, this approach raises concerns about factual consistency and the reliability of context grounding.

Siriwardhana et al. \cite{c2}. proposed a more dynamic RAG system by fine-tuning both the retriever and generator, and by updating the retrieval index continuously. Their system performs well on tasks like COVID-19 and multi-turn QA, but it also depends heavily on large-scale corpora and synthetic data. Their assumption is that better domain alignment comes from large-scale retraining, which can be resource-intensive and prone to propagating noise from web content.

Another notable related project is the Pet-Care-RAG-Application by Wufengyou \cite{c3}, which applies a RAG pipeline to pet care queries. While their work demonstrates a functional RAG setup, it relies heavily on a Kaggle dataset consisting of 583 question-answer pairs, likely sourced from unverified online platforms. This raises concerns about factual accuracy, consistency, and the presence of misinformation, which we think is problematic in sensitive domains like pet care where users seek reliable and health information.

From these previous work, our work shares the common goal of enhancing LLM performance in narrow domains but diverges in several key ways. First, we prioritize source credibility over scale by using a static, manually curated dataset sourced from trusted dog care and veterinary websites. Second, unlike prior work that focuses solely on RAG pipelines, we also explore LoRA-based fine-tuning as a lightweight, parameter-efficient alternative. This dual approach allows for a comparative evaluation of retrieval-based versus parameter-based specialization, offering insights not addressed in the above works. Also, our focus on dog care is highly practical but often overlooked, as there are few or no past work on dog care guidance.

\section{Methods}

\subsection{Data Collection}

\begin{figure}
    \centering
    \includegraphics[width=1\linewidth]{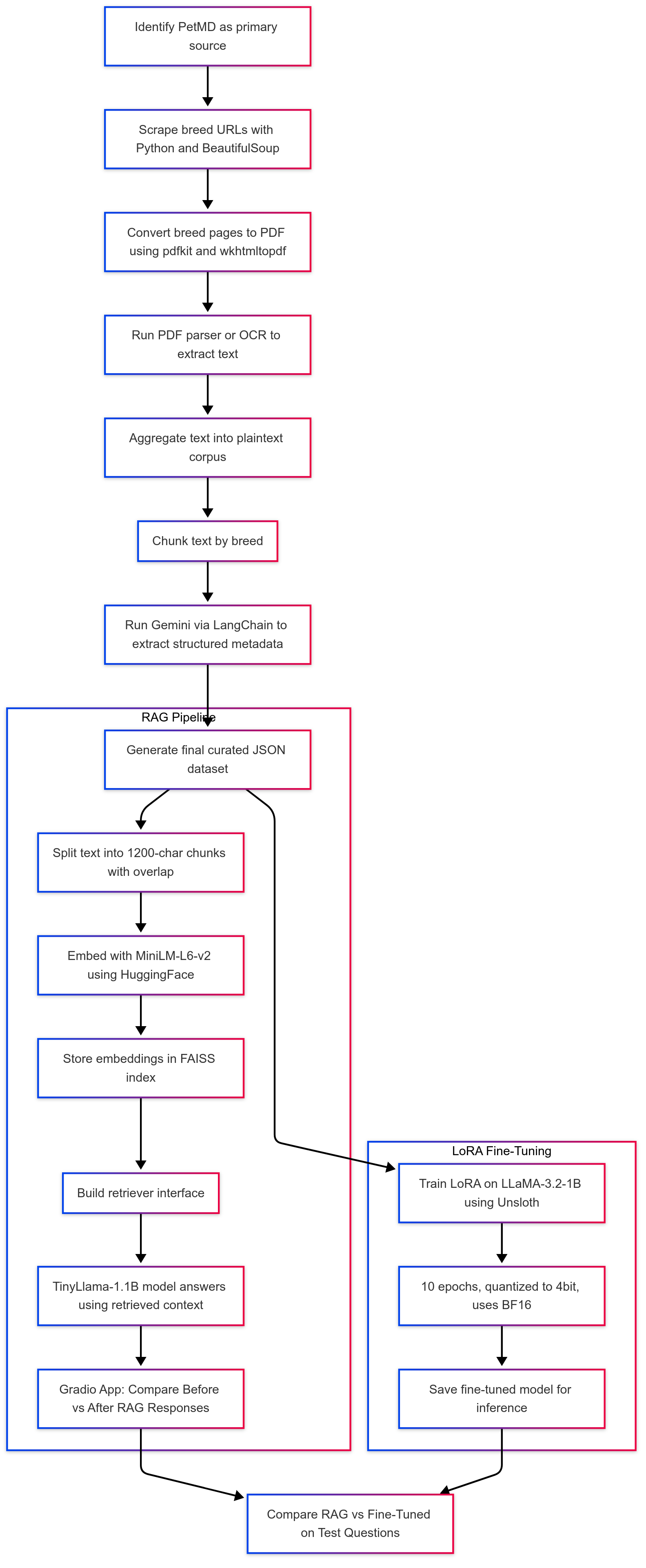}
    \caption{Architecture of the RAG and LoRA pipelines, both built from the same curated dog care dataset and evaluated on shared questions}
    \label{fig:enter-label}
\end{figure}

We first identifying reputable pet care websites, which has all the websites we came across. we chose PetMD \cite{c4} as our primary source for dog breed information, because it consists of a lot of information on all (260) breeds of dogs verified by veterinary sources. We felt this is enough rather than going to collect from various source to avoid duplications. We extracted specific dog breed details from the PetMD website. To do this, we use the requests library from Python to fetch the contents of the PetMD dog breeds page and BeautifulSoup to parse the HTML. We iterated through all anchor tags on the page and filtered out the links that matched the breed-specific URL pattern. For any relative URLs, we prepended the base domain to construct full URLs. We stored these unique links in a set and then printed them in alphabetical order.

\subsection{Data Processing}

With the list of breed URLs, we proceed to capture each page’s contents for offline access and future processing. We installed pdfkit and ensured that the wkhtmltopdf binary was available in the environment, enabling us to convert web pages to PDF format. We then looped through each breed URL, generated a user-friendly name for each breed by cleaning and formatting the last part of the URL, and converted the webpage to a PDF. These files were saved locally, and to be considerate of the PetMD server, we included a delay of 10 seconds between each request. This process allowed us to build a local archive of breed profiles, each saved as a PDF with accurate and consistent formatting. Once all breed pages were saved as PDFs, we packaged the entire folder into a ZIP archive and triggered the download to our local machine.

\subsection{Data Structuring}

Given the nature of HTML-to-PDF conversion, we anticipated a mixture of structured and rasterized content. To handle this, we use a  PDF parser capable of accessing embedded text layers in the documents. It was used as the primary extraction method, particularly effective for PDFs generated directly from HTML sources with minimal styling obfuscation. In cases where the parser returned empty outputs because when PDFs were rasterized and image-based, we converted each page of the PDF into high-resolution images and used a deep-learning-enhanced OCR engine to extract the text. Although computationally more expensive, this ensured no information was lost due to formatting anomalies. Once all individual documents were processed, the outputs were aggregated into a single plaintext corpus. Each section of the consolidated file was demarcated using the original PDF filenames as headers, which allowed manual inspection.

\subsection{Dataset Curation}

To prepare the data for LLM inference, our script first reads the entire file and splits the content into individual chunks based on header. These chunks represent unique breed entries. Each chunk is then processed through a structured data extraction pipeline. We constructed a prompt instructing the Gemini model to extract comprehensive breed-level metadata. 

The prompt specifies multiple categories of interest, including physical characteristics, lifespan, temperament, training tips, grooming requirements, health issues, feeding considerations, living environment, and any relevant historical or cultural notes. We  encourage the model to include unique or granular information when available and to return a JSON array of objects conforming to the specified schema. If any fields are not found in the text, they are to be marked as null. After invoking the model through LangChain \cite{c5}, the raw output is cleaned to remove any markdown formatting artifacts and is parsed into a Python object. Given the potential for API failures or inconsistent outputs, we implemented a retry mechanism that attempts to process each chunk up to five times, with exponential backoff delays between retries. If a chunk consistently fails, it is skipped, and a log message is recorded for post-processing analysis. The final output is stored in a global list, which accumulates valid model responses across all chunks. The aggregation logic is designed to handle both individual JSON objects and lists of objects, depending on the structure returned by the LLM. Once all chunks are processed, we aggregated the dataset into a file.

\subsection{RAG Framework}

Given the static nature of our dataset, we structured it for use in a retrieval-augmented generation (RAG) pipeline. We first divided the curated dog breed corpus into smaller chunks using a character-based splitter with a chunk size of 1200 characters and a 200-character overlap. This produced 1,229 chunks, balancing granularity with context preservation.

For embedding, we used the HuggingFace \texttt{sentence-transformers/all-MiniLM-L6-v2} model. The text chunks were processed in batches and converted into dense vector representations. Once embeddings were generated, they were stored in a FAISS vector index for fast similarity search.

After building and caching the vector store, we constructed a retriever interface to fetch relevant documents during query time. For the response generation component, we fine-tuned our system using TinyLlama-1.1B-Chat-v1.0, a lightweight causal language model. To make the model fit within Colab GPU memory, we applied 4-bit quantization using BitsAndBytes with \texttt{NF4} quantization and \texttt{float16} computation.

\subsection{LLaMA Fine-Tuning}

While RAG improved grounding, it required maintaining a retriever and external context. We explored LoRA fine-tuning to embed domain knowledge directly into model weights for more efficient inference.

To specialize the Unsloth/LLaMA-3.2-1B-Instruct model for the domain of breed-specific dog care, we applied LoRA, a parameter-efficient fine-tuning technique that allows large models to be adapted with minimal computational overhead. The model was loaded using 4-bit quantization via the BitsAndBytes configuration to reduce memory usage, with the compute data type set to \texttt{torch.bfloat16} to take full advantage of the RTX 4080 GPU’s performance capabilities.

For the LoRA setup, we used a configuration with a rank of 16 and a scaling factor (\texttt{lora\_alpha}) of 32, targeting the \texttt{q\_proj} and \texttt{v\_proj} modules within the transformer blocks. This resulted in only 0.22\% of the model’s total parameters being trainable, approximately 1.7 million out of 776 million, which enabled efficient training without compromising model adaptability.

The training was conducted over 10 epochs, using a batch size of 4 with gradient accumulation steps of 4 to achieve an effective batch size of 16. We set the learning rate to \texttt{5e-6} and used a cosine learning rate scheduler to manage decay. Mixed precision training with BF16 was enabled to optimize memory usage and speed. Evaluation was performed at the end of each epoch to monitor validation loss and ensure training stability. The Hugging Face \texttt{Trainer} was used to manage the training loop. Upon completion, the fine-tuned model was saved to \texttt{./fine\_tuned\_model\_dogcare}. Thanks to the efficiency of LoRA and the computing power of the RTX 4080, the entire fine-tuning process completed in approximately 10 to 20 minutes.

During development, we initially attempted to use the larger Meta-LLaMA-3-8B model, but encountered compatibility issues with the \texttt{xformers} library due to the GPU’s compute capability. As a result, we opted for the smaller 1B variant, which proved more stable and resource-efficient. In early fine-tuning attempts using Unsloth’s memory-optimized implementation, we ran into a \texttt{memory\_efficient\_attention\_backward} error. This led to the removal of Unsloth from the pipeline and a reconfiguration of the attention mechanism to rely on PyTorch’s default implementation. Additionally, we discovered a regression in one of the final test runs, caused by an incorrect model state. This was resolved by re-running the full 10-epoch training loop after debugging environment variables, verifying model file integrity, and correcting API-related changes. These iterative refinements ensured a stable, reproducible fine-tuning process.

\section {Results}

\subsection{RAG Evaluation}

We evaluated our system by comparing the performance of TinyLlama-1.1B-Chat-v1.0 both before and after integrating retrieval augmentation (RAG). To visually demonstrate the impact, we built an interactive web application using Gradio\cite{c8}. Gradio was selected for its simplicity, flexibility, and suitability for rapid prototyping of machine learning interfaces. It allows developers to wrap Python functions into shareable, browser-based UIs with minimal code overhead. In our case, it enabled real-time, side-by-side comparison of model outputs with just a few lines of code, making it an ideal platform for interactive evaluation. The ability to create clean and user-friendly input/output components—such as labeled textboxes—helped us clearly communicate differences in model performance to both technical and non-technical users. Additionally, Gradio supports public sharing (share=True) without requiring a deployment server, which was particularly useful for demonstration and feedback during class presentations and testing phases.

We posed a set of 15 expert-verified dog care questions to two variants of the system:

\begin{itemize}
    \item \textbf{Before RAG:} TinyLlama responded directly without access to any external information.
    \item \textbf{After RAG:} TinyLlama generated responses using breed-specific context retrieved from our FAISS knowledge base.
\end{itemize}

In the baseline (\textit{Before RAG}) setting, TinyLlama often produced generic, repetitive, or hallucinated responses. For example, when asked, \textit{“How should I care for a Curly-Coated Retriever’s coat?”}, the direct model redundantly suggested brushing the dog’s coat with a soft-bristled brush in nearly identical steps, without acknowledging that overbrushing can damage the curl structure, a critical breed-specific detail.In contrast, the \textit{After RAG} response included structured and actionable advice sourced from verified context. It recommended brushing with a slicker brush or comb to avoid tangles, checking for signs of skin issues, nail trimming with appropriate length, and noted that excessive brushing can damage the coat, highlighting a far better alignment with veterinary guidelines.

\begin{figure}[H]
    \centering
    \includegraphics[width=1\linewidth]{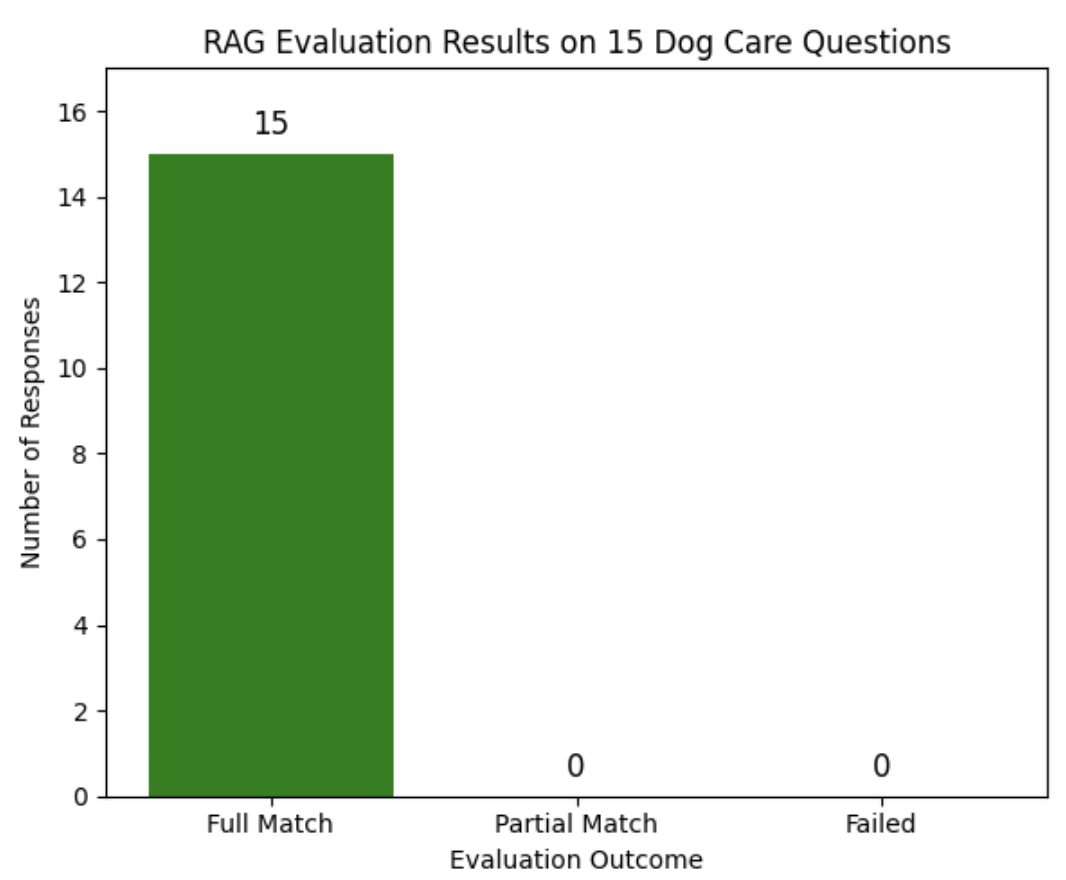}
    \caption{RAG Evaluation Results on 15 Dog Care Questions}
    \label{fig:enter-label}
\end{figure}

To systematically evaluate these improvements, each generated response was manually compared against expert ground truth. Outputs were labeled as full match, partial match, or no match. The RAG-augmented system achieved a perfect record, with 100\% of responses classified as full matches across all 15 queries.

The strongest improvements appeared in responses requiring detailed reasoning about grooming, training, and health conditions, domains where breed-specific nuance is essential. These results affirm that retrieval grounding significantly enhances the factuality, specificity, and reliability of TinyLlama’s responses beyond what its pretraining alone can provide.

\subsection{Fine Tuning Evaluation}

To evaluate the impact of fine-tuning, we first established a baseline using the pre-trained \texttt{unsloth/Llama-3.2-1B-Instruct} model. We tested it on a set of 15 carefully curated dog care questions that spanned various topics including training, diet, grooming, health, and behavior. The baseline performance revealed clear limitations: many responses were off-topic or lacked the domain specificity required for practical, breed-aware advice. For example, the model frequently offered training tips when asked about grooming or returned generic suggestions with no mention of breed, age, or condition-specific nuances. This helped us establish a reliable benchmark to measure the effectiveness of our LoRA-based fine-tuning.

\begin{figure} [H]
    \centering
    \includegraphics[width=1\linewidth]{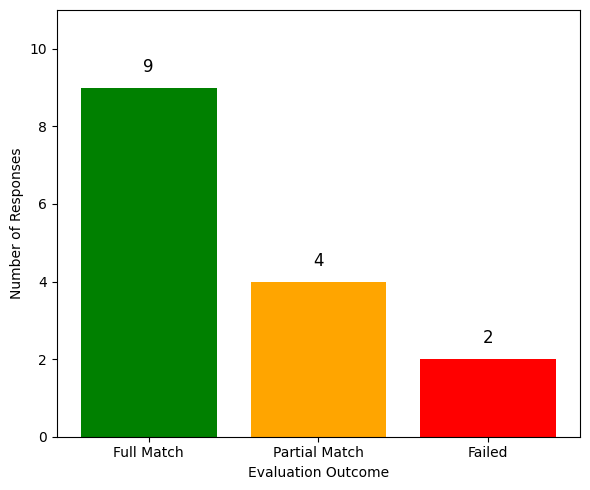}
    \caption{Fine-Tuning Evaluation Results on 15 Dog Care Questions}
    \label{fig:enter-label}
\end{figure}

After completing the fine-tuning process, we re-evaluated the updated model on the same set of 15 questions. The outputs were collected using a corrected testing script that saved the predictions to a CSV file. Each fine-tuned output was then compared to a ground truth response curated from expert sources. The evaluation was conducted manually, classifying each result as a full improvement, partial improvement, or no improvement based on relevance, accuracy, and alignment with expert recommendations.

The results showed a marked improvement in the model’s responses. Out of the 15 test queries, 9 (60\%) showed full improvement, 4 (26.7\%) showed partial improvement, and only 2 (13.3\%) remained unimproved, yielding an overall improvement rate of 87\%. The most notable improvements were seen in training-related questions (1, 2, 3), where the model shifted from vague or unrelated suggestions to well-aligned responses emphasizing positive reinforcement strategies. Similarly, in socialization and diet-related questions (4, 5, 6, 7), the model exhibited a clearer understanding of the context, offering more relevant and concise advice.

Grooming (question 9) and health (questions 12 and 13) showed partial improvements, the model recognized the right topics but sometimes failed to mention critical breed-specific practices, such as appropriate bathing frequency or detailed symptoms. For example, while it recommended brushing for grooming, it omitted the detail that certain breeds should only be bathed a few times per year to maintain coat health.

However, two queries (10 and 14) failed to improve. In the case of the Curly-Coated Retriever’s grooming routine, the model incorrectly suggested brushing when coat care requires more specific strategies. Similarly, it misattributed the cause of seizures in question 14, pointing to general mutations rather than the more accurate explanation found in expert literature.

These shortcomings indicate that the 644-example dataset, while rich, may not yet cover the full breadth of breed-specific edge cases. Additional training for additional epochs could help address these residual gaps and push performance further.

\subsection{Comparative Analysis}

\begin{figure}[H]
    \centering
    \includegraphics[width=1\linewidth]{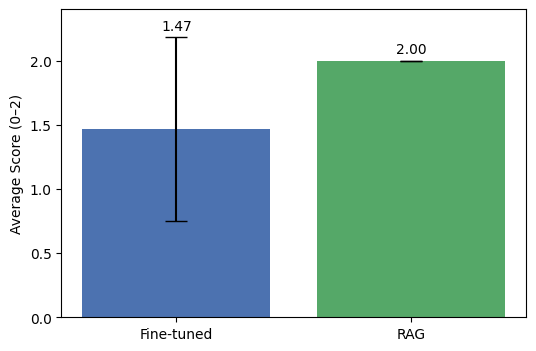}
    \caption{Comparison of Fine‑tuned vs. RAG Model Performance}
    \label{fig:enter-label}
\end{figure}

We computed the mean and standard deviation of the performance scores for 15 test questions, assigning numerical values to qualitative ratings (Full = 2, Partial = 1, None = 0). The fine-tuned model yielded an average score of 1.47 ($\sigma = 0.72$), reflecting a mix of strong and underwhelming responses. In contrast, the RAG model achieved a perfect score of 2.0 with zero variance, indicating it consistently produced fully correct answers. This underscores the value of retrieval-based grounding for maintaining both accuracy and reliability in domain-specific tasks.

In our comparative analysis, we first ran a Welch’s $t$‑test, which does not assume equal variances between the fine‑tuned and RAG scores. The result ($t = -2.779$, $p = 0.0148$) indicates a statistically significant difference, with the RAG system outperforming the fine‑tuned model. To confirm this finding under a non‑parametric framework, we also performed a Mann–Whitney U test, yielding $U = 67.5$ ($p = 0.0081$). Both tests reject the null hypothesis ($p < 0.05$), demonstrating that the improvements seen with RAG are robust across parametric and non‑parametric analyses. 

Additionally, we treated the data as paired observations and conducted a Wilcoxon signed-rank test. The test yielded $W = 0.0$ ($p = 0.0231$), again indicating a statistically significant improvement from the fine-tuned model to RAG. To further quantify this improvement, we computed Cohen’s $d$ for paired samples, resulting in a moderate-to-large effect size of $d = 0.72$. Finally, we bootstrapped the mean difference between the two models, which produced a 95\% confidence interval of $(0.20,\ 0.93)$ for the performance gain, further reinforcing the practical significance of the RAG system’s improvements.

\begin{figure}[H]
    \centering
    \includegraphics[width=1\linewidth]{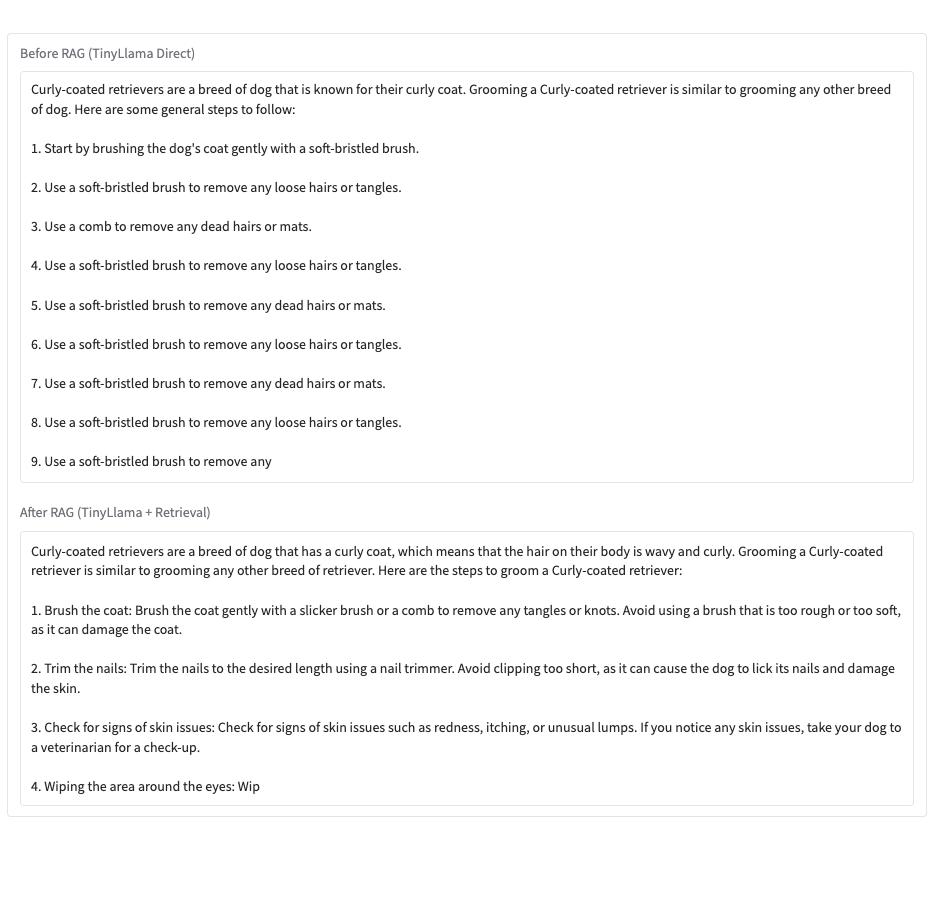}
    \caption{RAG Before and After Comparison}
    \label{fig:enter-label}
\end{figure}

\begin{figure}[H]
    \centering
    \includegraphics[width=1\linewidth]{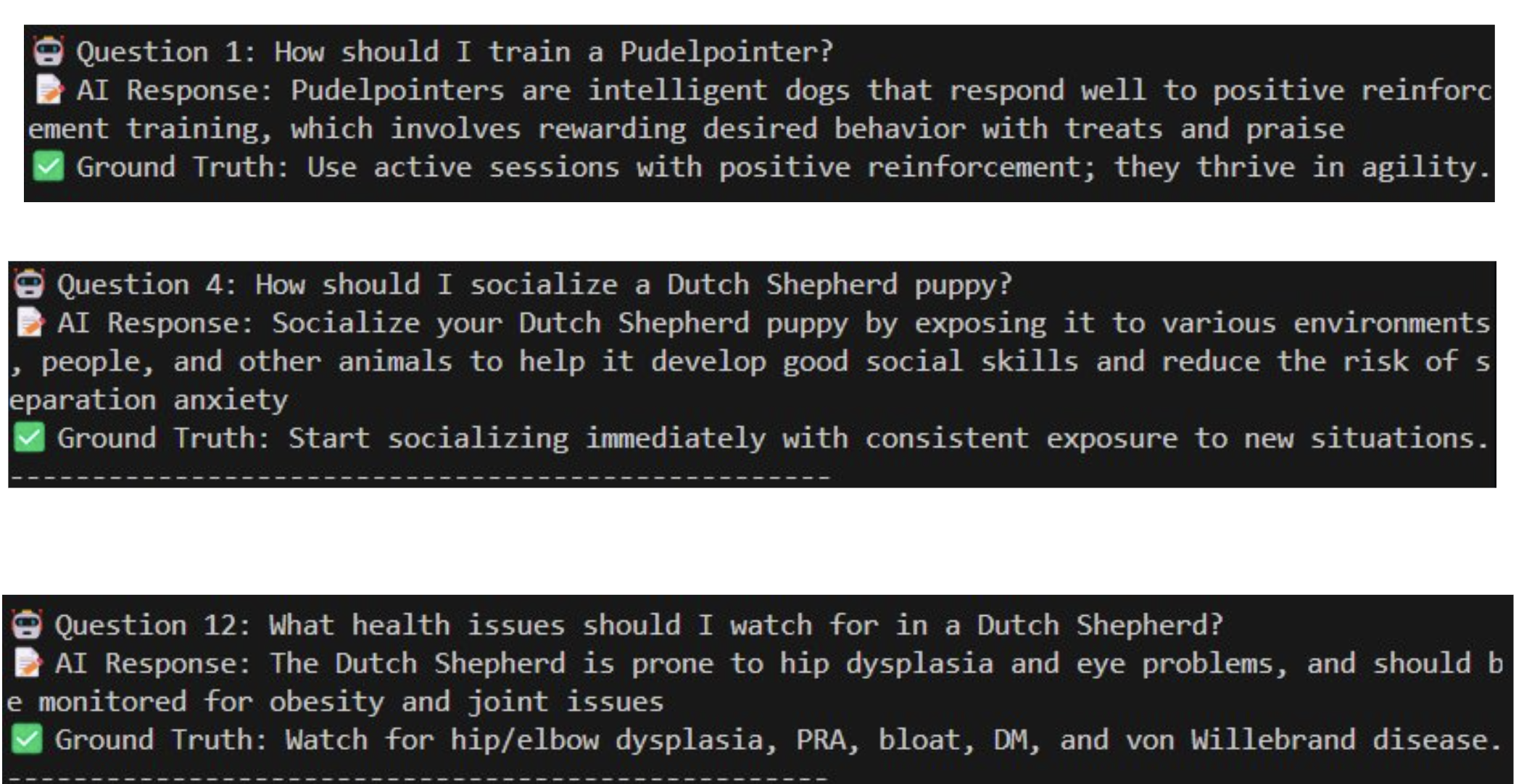}
    \caption{LoRA fine tuning}
    \label{fig:enter-label}
\end{figure}

\section{Discussion and Future Work}

\subsection{5.1 Revisiting the Research Questions}

\textbf{RQ1: Can domain grounding improve the factual accuracy and reliability of pet care advice provided by LLMs?}

Our results provide strong evidence that domain grounding significantly enhances the factual accuracy and reliability of language model outputs in the pet care domain. The RAG system consistently delivered responses that were better aligned with expert-verified ground truth answers, achieving a 100\% acceptable response rate and a perfect average score of 2.0. This performance, coupled with zero variance across the 15 test queries, demonstrates that domain-grounded retrieval is highly effective in reducing generic, vague, or inaccurate content in LLM responses.

\vspace{1em}  

\textbf{RQ2: What is the performance gap between generic responses from baseline LLMs and enhanced outputs derived from curated RAG and fine-tuned approaches?}

Both the RAG and fine-tuned models showed substantial improvements over baseline outputs, which were often generic or off-topic. The fine-tuned model reached an 87\% improvement rate and a mean score of 1.47, while the RAG system achieved full marks across all metrics. Statistical analysis confirmed the performance gap, with Welch’s t-test ($p = 0.0148$), Mann–Whitney U ($p = 0.0081$), and Wilcoxon signed-rank test ($p = 0.0231$) all indicating significant differences. Cohen’s $d = 0.72$ further emphasized the practical significance of RAG’s advantage. These findings suggest that while both strategies improve performance, RAG offers the most consistent and reliable enhancement.

\subsection{Limitations}
Despite promising results, several limitations constrain the generalizability of our findings. First, our evaluation was based on a relatively small dataset of 15 curated dog care questions. Although carefully selected, this test set does not fully capture the diversity of real-world pet care queries, and a larger-scale evaluation would be necessary to establish statistical robustness.

Second, we were limited to fine-tuning a 1B parameter model due to hardware constraints (RTX 4080). This restricted our ability to explore whether larger models (e.g., 7B or 13B) would respond better to LoRA fine-tuning. Similarly, our RAG setup relied on a static dataset and did not incorporate dynamic retrieval or retriever fine-tuning, which could further enhance response quality.

Additionally, neither system was evaluated in a live user-facing setting, and as such, we lack feedback on perceived helpfulness or satisfaction from real users. This limits our understanding of the practical utility of these systems in interactive scenarios.

\subsection{Lessons Learned}
Several technical and methodological insights emerged throughout this project. We encountered and overcame compatibility issues between certain libraries (e.g., xformers and Unsloth), highlighting the importance of aligning software stacks with available hardware. The iterative debugging of fine-tuning regressions also emphasized the need for reproducibility and validation checkpoints throughout model development. On the methodological side, we learned that carefully curating a dataset from authoritative sources often yields better factual accuracy than relying on large-scale, noisy data.

\subsection{Future Work}
Future iterations of this work could explore hybrid architectures that combine the strengths of both approaches—e.g., a fine-tuned model that also uses lightweight retrieval as a fallback. Furthermore, real-time user studies or feedback collection would provide valuable insights into the trustworthiness, usability, and real-world impact of these tools. Finally, retriever fine-tuning and integration with multimodal inputs (e.g., images of pets) represent promising directions for broadening the system’s capabilities.



\begin{thebibliography}{99}

\bibitem{c1} H. Sun, Y. Wang, and S. Zhang, "Retrieval-Augmented Generation for Domain-Specific Question Answering: A Case Study on Pittsburgh and CMU," Carnegie Mellon University, Pittsburgh, PA, USA. arxiv.org/pdf/2411.13691 [Accessed: 11-April-2025].
\bibitem{c2} S. Siriwardhana, R. Weerasekera, E. Wen, T. Kaluarachchi, R. Rana, and S. Nanayakkara, "Improving the Domain Adaptation of Retrieval Augmented Generation (RAG) Models for Open Domain Question Answering," arXiv preprint arXiv:2210.02627, Oct. 2022. [Accessed: 11-April-2025]
\bibitem{c3} F. Wu, "Pet Care RAG Application," GitHub repository, 2024. [Online]. Available: github.com/wufengyou/Pet-Care-RAG-Application. [Accessed: 20-Feb-2025].
\bibitem{c4} PetMD, "The Best Pet Health \& Care Advice from Real Vets," [Online]. Available: https://www.petmd.com/ [Accessed: Apr. 11, 2025].
\bibitem{c5} LangChain, "LangChain: Applications That Can Reason," [Online]. Available: https://www.langchain.com/. [Accessed: Apr. 11, 2025].
\bibitem{c6} E. Y. Çalık and T. R. Akkuş, "Enhancing Human-Like Responses in Large Language Models," \emph{arXiv preprint arXiv:2501.05032}, Jan. 2025. [Online]. Available: https://arxiv.org/pdf/2501.05032.
\bibitem{c7} C.-C. Hung, W. B. Rim, L. Frost, L. Bruckner, and C. Lawrence, "Walking a Tightrope—Evaluating Large Language Models in High-Risk Domains," \emph{arXiv preprint arXiv:2311.14966}, Nov. 2023. [Online]. Available: https://arxiv.org/abs/2311.14966.

\bibitem{c8}
A. Abid, M. Chaudhary, and J. Zou, “Gradio: Hassle-Free Sharing and Testing of ML Models in the Wild,” *arXiv preprint arXiv:1906.02569*, 2019. [Online]. Available: https://arxiv.org/abs/1906.02569




\end{thebibliography}
\end{document}